# Transfer Learning in Deep Learning Models for Building Load Forecasting: Case of Limited Data

Menna Nawar[1], Moustafa Shomer[1], Samy Faddel[2], and Huangjie Gong[2]
[1]Alexandria Higher Institute of Engineering and Technology
[2]US Research Center, ABB Inc.
Corresponding author: samy.faddel-mohamed@us.abb.com

*Abstract*—Precise load forecasting in buildings could increase the bill savings potential and facilitate optimized strategies for power generation planning. With the rapid evolution of computer science, data-driven techniques, in particular the Deep Learning models, have become a promising solution for the load forecasting problem. These models have showed accurate forecasting results; however, they need abundance amount of historical data to maintain the performance. Considering the new buildings and buildings with low resolution measuring equipment, it is difficult to get enough historical data from them, leading to poor forecasting performance. In order to adapt Deep Learning models for buildings with limited and scarce data, this paper proposes a Building-to-Building Transfer Learning framework to overcome the problem and enhance the performance of Deep Learning models. The transfer learning approach was applied to a new technique known as Transformer model due to its efficacy in capturing data trends. The performance of the algorithm was tested on a large commercial building with limited data. The result showed that the proposed approach improved the forecasting accuracy by 56.8% compared to the case of conventional deep learning where training from scratch is used. The paper also compared the proposed Transformer model to other sequential deep learning models such as Long-short Term Memory (LSTM) and Recurrent Neural Network (RNN). The accuracy of the transformer model outperformed other models by reducing the root mean square error to 0.009, compared to LSTM with 0.011 and RNN with 0.051.

## I. INTRODUCTION

As a result of population growth and ongoing technological and economic developments, the number of commercial buildings has increased by 6% from 2012 to 2018 [1]. Newer buildings seem to be bigger in size and have modern electronic devices and more sophisticated equipment, which led to more power consumption and energy bills. In 2018, the U.S. has consumed more energy than ever before recording 101.3 quadrillion Btu, up 4% from 2017 [2]. In the following year 2019, U.S. has represented 17% of the energy consumption out of the whole world consumption [3]. Therefore, it is essential for respective parties involved in energy management such as governments and companies to improve the power consumption efficiency.

Load-forecasting of commercial buildings plays a crucial role in increasing energy schedule efficiency, and has become a serious topic that promoted a substantial degree of research. The principal obstacles arise from the fact that there is not a specific factor that controls the energy consumption in commercial buildings, instead there are multiple factors affecting the change of the consumed power. Those factors could affect the forecasting either in a direct way or indirect way, including weather conditions, building size, building type, electronic equipment and human activities, etc. [4].

Since forecasting is a key in improving the energy efficiency issue in commercial buildings, it is necessary to understand different types of load-forecasting. According to the spectrum of time intervals, load forecasting problems are classified into three types: short-term load forecasting (STLF) [5], medium-term load forecasting (MTLF) and long-term load forecasting (LTLF) [6]. This paper focuses on MTLF given the adopted time interval of load forecasting. MTLF has been an active research area which is highly discussed in the literature since it delivers valuable information for both planning and operation [7]-[10]. MTLF can be assumed from one week to several months and has a goal of making a systematically effective operational plan and scheduling energy for both power generation plants and distribution utilities.

In the past decades, it has been observed that improving the accuracy of energy forecasting has become an active issue. It was discussed in many studies using conventional statistical techniques and Artificial Intelligence (AI) based approaches. Traditional statistical techniques usually involve auto-regressive integrated moving average (ARIMA), linear regression, and exponential smoothing [11], [12]. Statistical methods are fast and easy to apply because they rely on linear functions to process the relationships between the historical and forecasted data. As the time-series load forecasting is a non-linear problem, these models are not always forecasting the load satisfactorily. This issue motivated the development of AI-based models as they are non-linear models which apply non-linear functions to forecast the load demand.

Driven by the suitability and efficiency of AI techniques, they have become common in the literature to solve arduous load forecasting problems [13], [14]. AI-models can be categorized as Machine Learning and Deep Learning models.

In terms of Machine Learning models, since these models have advantages such as simplicity and high computation speed, they are highly discussed in the literature [15], [16]. However, Deep Learning models, have proved to be more accurate in energy forecasting tasks [17]. AI-models depend on a large amount of historical data to predict power consumption [18], [19]. In [20], the authors used Deep Learning to predict humidity, air temperatures and energy behavior of buildings.

The algorithm was trained on a large amount of data. Relying on a large-scale data to predict energy consumption was also used in [21], where the authors used a hybrid Deep Learning model that consists of Long Short-Term Memory (LSTM) and Recurrent Neural Network (RNN) in LTLF and trained their model on six years of data points from 2004 to 2010. In [22], the authors adapted a developed deep neural network (DNN) for heating energy consumption forecasting. Although, the model had an acceptable performance when it was compared to simulation model from EnergyPlus, the authors did not consider comparing their model to other popular Deep Learning models that are suitable for time-series forecasting.

Transformer model is considered one of the newly suggested models in the literature that was mentioned to have a great potential. In [23], the authors trained the Transformer model on 87,648 sampling data point before applying it in STLF, which makes it impractical for new buildings or those that do not collect data at a frequent basis. In [24], the authors explored transfer of knowledge from another domain, then applied this knowledge in two machine learning models (Random Forrest and Feed Forward Network). In [25], Transformer model was trained on a very huge-scale of data (19 years of data points) for wind speed forecasting, which clearly proves that Transformer model is considered one of the most data-hungry models. One of the most recent studies [26], The authors modeled Multi-Layer Perceptron (MLP) and Long Short-Term Memory (LSTM) on a medium-sized office building with data scarcity to predict the building thermal dynamics. They applied both traditional Machine Learning and Transfer Learning. The results of deep analysis leveraged using Transfer Learning and lead to higher accuracy. It is noteworthy that none of the previous work considered the use of Transformer model with Transfer learning to tackle the case of limited date.

This paper proposes a medium-term load forecasting methodology that can be used for buildings with scarce historical data. The methodology is based on taking the knowledge and information learnt from buildings' data in the same domain and transferring it to the targeted building. The paper also applies the framework of knowledge transfer to the Transformer deep learning model that is known for its ability in capturing trends and relations. To the best of our knowledge, this is the first time to apply transfer learning to the Transformer model in energy forecasting domain. Finally, the paper compares the proposed methodology to common deep learning techniques such as RNN and LSTM.

The paper is organized as follows: Section 2 introduces the transfer learning and background concepts concerning the transfer learning technique. Section 3 presents deep learning models and discusses the specific models used in this paper. Furthermore, Section 4 shows a case study used to evaluate the proposed method and its results. Finally, our conclusion and future work will be discussed in section 5.

## II. TRANSFER LEARNING TECHNIQUE

### A. Sequential Deep Learning Models

Deep Learning (DL) has become progressively popular in the field of time-series forecasting. The building-unit of any DL model is neural network. The neural network receives inputs, which can be text, image, video, or time-series data like this paper. Then, the input is processed in one or more hidden layer by using weights. These weights can be adjusted during the training process. Finally, the model comes up with the prediction in the output layer.

Sequential DL models are the models that use temporal data as inputs. Time-series data are form of sequence data. The main advantage of sequence models is assuming that all inputs and outputs are dependent of each other. Therefore, these models can work efficiently in energy forecasting since the previous inputs in time-series data are inextricably significant in predicting the energy-consumption output. Recurrent Neural Network (RNN) and Long Short-Term Memory (LSTM) are popular sequential models used in energy forecasting using time-series data as inputs, while Transformer model is rarely used in this domain. Noteworthy, the three models can capture the patterns and trends, especially the Transformer. The architecture of each model will be discussed in detail in section III, and based on the patterns and trends, the models can predict accurately the energy consumption value.

### B. Building-to-Building Transfer Learning

Transfer Learning (TL) is a promising approach that can help in enhancing the model performance by applying the prior knowledge gained from a source task to a similar or different target task. This approach was inspired by the way humans learn new things. Humans have always benefited from their transfer of knowledge and its usage in other different areas. From computer science point of view, the aim of TL is to leverage a pre-trained model's knowledge, then applying this knowledge in performing another task. Fig. 1 illustrates the difference between traditional learning in part (a) and Transfer Learning in part (b). It shows that Transfer Learning depends on transferring the knowledge to a pre-trained model. Hence the complexity of the training process will be much easier.

Considering the load forecasting problem, it is customary using massive amount of historical data to train an AI-based model to predict the energy consumption. Thus, relying on a large data has become unavoidable, which is impracticable in the case of newly constructed buildings. Unlike TL case, it is available to use scarce historical data to train the DL model as long as it is possible to take advantage of previous knowledge gained from an older task or another source task.

Given that the DL models are always data-hungry, the main advantage of TL in these models is that it makes the use of small amount of data for training not only possible, but also efficient. Nevertheless, other advantages of TL are faster training, better model-initializing and higher learning rate for training. From a theoretical perspective of TL, the source (first task) and target tasks (other tasks) could be from the same or different domains. When it comes to load forecasting field, it is more popular in the literature to apply TL between two different domains. This paper demonstrates for the first-time applying building-to-building Transfer Learning in MTLF. The pre-trained models used in this study are initially trained for a load forecasting task for an old building. Then, the prior knowledge gained from training on previous data, is used to enhance the accuracy of MTLF for new or other commercial buildings with scarce data. Briefly, TL in this paper is mainly used to acquire the knowledge from some old buildings to improve modelling/forecasting efficiency on new buildings, with limited data.

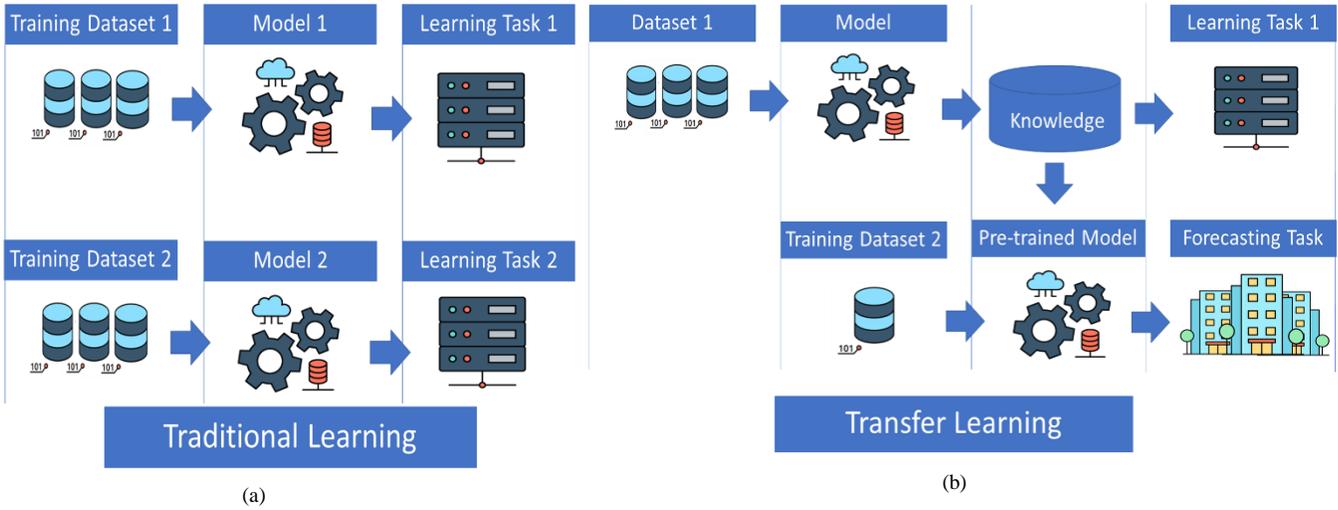

Fig. 1 The difference between traditional learning (a) and transfer learning (b).

## III. DEEP LEARNING MODELS

### A. RNN and LSTM Models

RNNs have the same general principle of ANNs. While traditional ANNs can only deal with the input data individually and process the relationships between inputs and outputs without linking each output to the preceding. The main difference that RNNs can address the sequential data by using internal memory and assume that every output is related to the previous state. Once the output of the RNN is generated, it is copied and returned to the system as an input, hence the name "Recurrent" of the RNN comes from. Fig. 2 illustrates the structure of RNN, where X and O indicate the input and the output through the time t. Hidden layers are represented by h. U and V denote the weight matrices between inputs and hidden layers, hidden layers and outputs. W represents the weight metric between hidden layers in several time stamps.

Although RNN is capable of processing sequential data, it fails to extract long-term information after many repeated iterations, because RNN suffers from vanishing gradient problem (this means that the internal memory of RNN cannot keep tracking of the gradient after multiple loops). To overcome this problem, LSTM networks were proposed in [27]. LSTMs are considered modified RNNs, however LSTMs are more effective in time-series forecasting for the long-term dependency than RNNs. The structure of LSTM is similar to RNN, with an addition hidden state, it is also called LSTM cell as shown in Fig. 3 (where c denotes the computations of weights).

This cell works as a separate memory for the network, and it is responsible for remembering the long-term information in LSTMs. This cell contains three gates; forget, input and output gate. The forget gate decides which state in the cell can be forgotten, as this information is no longer necessary for the next state prediction.

The input gate is responsible for adding or updating the internal cell with the new information. The output gate selects which part should be addressed as an output among all of the information in LSTM cell. The gates in LSTM are represented by the following equations:

$$i_t = \sigma(w_i[h_{t-1}, x_t] + b_i) \quad (1)$$

$$f_t = \sigma(w_f[h_{t-1}, x_t] + b_f) \quad (2)$$

$$o_t = \sigma(w_o[h_{t-1}, x_t] + b_o) \quad (3)$$

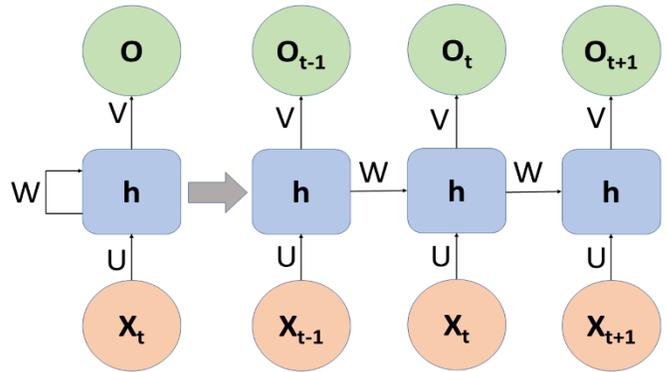

Fig. 2 Fundamental topology of RNN

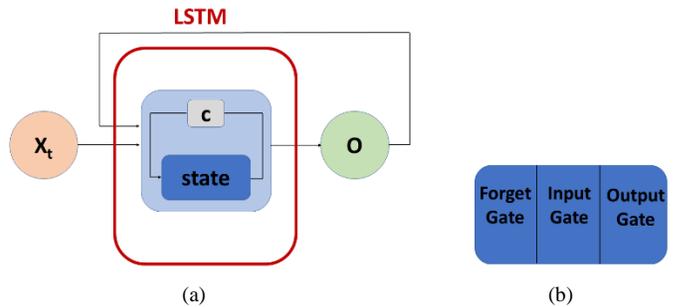

Fig. 3 Simple architecture of LSTM (a), the internal state or the cell of LSTM and its gates (b).

where: $i_t$ is the input gate, $f_t$ is the forget gate and $o_t$ is the output gate of the network. Sigmoid function is represented by $\sigma$, $w$ denotes the weight for the respective gates and $h_{t-1}$ is the previous output at timestep $t-1$. The current input is represented by $x_t$ and $b$ indicates the biases for the respective gates.

Although LSTM has overcome the obstacle of vanishing gradient in RNN, this model created other challenges. Since relying on the LSTM cell including the three gates, the information from previous steps has to go through a long sequence of computations. This problem leads to difficulty in training LSTM due to a very long gradient path. This recursion of the sequence will cause information loss eventually. As mentioned earlier, RNN and LSTM suffer from forgetting the information in the long-term, since both models process the inputs sequentially. Hence, the need for a new model addressing the inputs in parallel rather than sequentially.

*B. Attention mechanism and Transformer Model*

The solution to the long-term dependency problem was solved in 2014 and 2015 by [28], [29]. These pioneering papers proposed a technique called the Attention Mechanism. Attention Mechanism is the backbone of the Transformer model, illustrated in [ Fig. 4], Transformers are currently the state-of-the-art solution for many problems such as computer vision and natural language processing. It consists of two parts, encoder and decoder, where each block contains one or more attention layer followed by a multi-layer perceptron (MLP). Using this technique dramatically improved the quality of any sequence-related tasks and used later in the time-series forecasting domain. The attention allows the model to concentrate on only the important and relevant subsets in the long sequences of the input. Concretely, the model has to decide by learning on its own which information from the past steps are relevant for encoding the available input, and then takes the encoded information and decodes it to representative features that is used for forecasting.

The core of attention mechanism is assuming that the data source contains elements that can be represented as Key (K) and Value (V). Another element; Query (Q) is associated with the target data, where: KT ∈ R3x5 , V ∈ R5x3 and Q ∈ R4x3. There are three steps to apply the attention mechanism [27]. First, calculating the similarity score between the key vector and the query vector according to (4). Secondly in (5), converting the similarity score into weights, then arranging these weights in a probability distribution. Finally, (6) represents calculating the attention value by weighted summation of all the resulted coefficients. The reader is referred to [27] for more information.

$$similarity(QV, k_i) = \frac{QV \cdot k_i}{\|QV\| \cdot \|k_i\|} \qquad (4)$$

$$W_i = softmax(similarity_i) \quad for \; i = 1, 2, 3, \dots, n \qquad (5)$$

$$Attention\; Score(Q, K, V) = \sum_{i=1}^{n} w_i \cdot V \qquad (6)$$

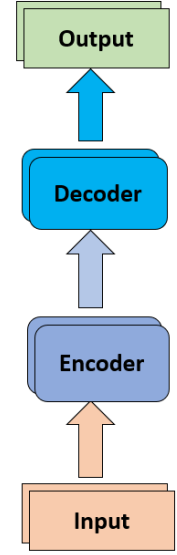

Fig. 4 The Architecture of the Transformer Model

In 2017, a new DL model, called Transformer, was proposed based on the attention mechanism [30]. Transformer was rapidly used in many different areas such as translation, image classification and time-series forecasting, and it overcame state-of-the-art models in these areas.

In comparison, the main advantage to the Transformer's architecture is that it allows the model to access the input data directly using parallel computation, not sequentially as the case of RNN and LSTM. Processing the inputs in parallel avoids the recursion and the iterations, which leads to reduction in the training time and the probability of losing information in the long dependencies. The Transformer does not depend on the previous hidden states to capture the patterns in order to predict the output. Instead, it processes a batch of input data as a one unit learning positional embeddings to encode the relationships between each observation and search for dependencies and patterns in the time-series data. Positional embedding is a technique that was introduced to replace recurrence by using weights that can encode the information related to a specific position of a certain input, then the transformer decodes the information and transforms it into prediction for the next time-step.

IV. CASE STUDY

This paper used hourly collected data from two buildings over time interval of one year starting from 1 January 2016 to 1 January 2017. The data were adopted from American Society of Heating, Refrigerating and Air-Conditioning Engineers (ASHRAE) [31]. The proposed models, Transformer, LSTM and RNN, were trained on a subset of that data, representing only 20% of the total samples, which counted up to 336 data samples. The time window used in the experimentations is 6 hours.

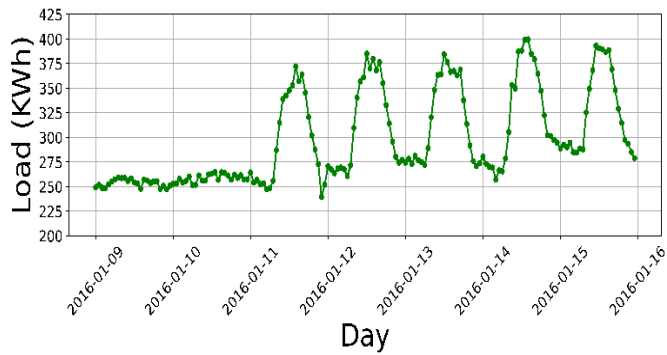

Fig. 5 The load consumption over one week for one building including weekend and workdays

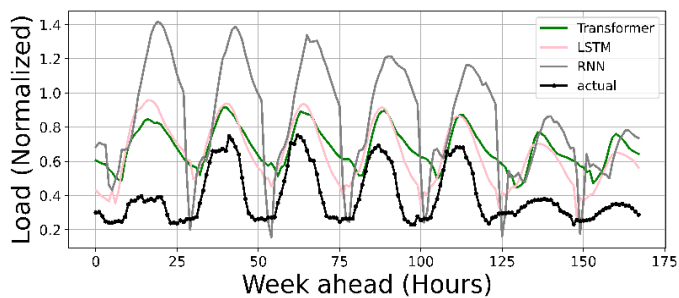

Fig. 6 The comparison of the three models

The dataset contained 14 features, including information about weather conditions like temperature, cloud coverage, precipitation depth and others. It also included buildings' metadata like buildings' sizes and primary usage. To improve the performance of the model, some engineered features were added like day-of-the-week, seasons, and day-of-the-month, to account for the vacations, holidays and seasonal trends. This paper focuses on educational buildings as they captured most of building's types by 38%. An example of the pattern of the load consumption over one week for one building is shown in Fig. 5 According to the floor count, the number of floors for the two buildings is equal to five floors, representing a large building type

Finally, the data used in the study had many features. Some of these features were representative and informative for the modeling process, while others where not relevant to the task or might harm the prediction. After analyzing the data, it was decided to go with the chosen featurism while ditching the rest.

An example for the neglected data is the sea level pressure as it is almost constant across all data points. Also, the wind direction as it does not affect the temperature, but it is an indication for the presence of the wind. This would force the model to learn this correlation, instead of learning about the desired task. As well as, engineering the features to give more information to the proposed model, such as the day, month, and hour. It is crucial to determine If the day is a weekend or not, because the usage of electricity highly depends on this feature. The model can also learn to combine information from multiple features together.

The results will first show how the transfer learning can improve the accuracy of the deep learning model, enabling forecasting in building with limited data. Then, the paper will compare the proposed transfer transformer model to other sequential models.

### A. Transfer-Learning Effect on the Transformer model

This section aims to see the effect of applying building-to-building Transfer Learning on improving the performance of the Transformer model which were trained on limited data. In theory, transfer learning takes the weights of the pre-trained model and uses them as a starting point to train from, which increases the speed and probability of better convergence to reach high performance. Since it is the first time to apply this type of Transfer Learning to the Transformer model in the domain of energy forecasting and train it on limited data to forecast the load demand, it is essential to evaluate the impact of applying Transfer Learning. Considering the case of large commercial buildings, after adapting building-to-building Transfer Learning the performance of the Transformer model has been improved by 56.8 % in terms of mean squared error (MSE), and almost 34 % in terms of root mean square error (RMSE) and mean absolute percentage error (MAPE) as shown in Table. I.

### B. Forecasting the Load of Large Buildings

To evaluate the performance of the proposed model compared to the traditional sequential models such as the RNN and the LSTM, the three models were trained for 15 epochs to prevent models' overfitting and make it easier to measure the performance of the models under the same training period. The same large building was used as the source of data for the three models. Also, transfer learning was applied to all of them to ensure apple to apple comparison. The results are presented in Fig. 6 and Table. II. The figure shows that the RNN model has the tendency to overestimate the load for both during heavy and light loading conditions compared to the other two models. The LSTM shows close performance to the transformer model though the transformer is better in following the trend, resulting in a better accuracy as given in Table. II. This result is because the Transformer is built for capturing the patterns, trends, and relations between data points. Regardless of the improvement in the accuracy, the three models do not seem to perform well under light loading conditions for the case of weekends as shown after hour 125 in Fig. 6. This will be investigated more in the future. However, typically load forecasting for weekends in commercial buildings is not of high importance since the load is very low and there are not any occupants in the building.

TABLE I. IMPACT OF TRANSFER LEARNING IN TRANSFORMER'S PERFORMANCE IN CASE OF LARGE BUILDINGS

| Type of Metric | Transformer model | |
| --- | --- | --- |
| | Before Transfer Learning | After Transfer Learning |
| MSE | 0.021 | 0.009 |
| RMSE | 0.146 | 0.096 |
| MAPE | 0.311 | 0.203 |

TABLE II. EVALUATION METRICES OF THE THREE MODELS IN LOAD-DEMAND FORECASTING OF LARGE BUILDINGS

| Model / Metric | Transformer | LSTM | RNN |
|---|---|---|---|
| MSE | 0.009 | 0.011 | 0.051 |
| RMSE | 0.096 | 0.106 | 0.227 |
| MAPE | 0.203 | 0.217 | 0.471 |

V. CONCLUSIONS

This paper proposed building-to-building transfer learning for energy load forecasting, which aims for enhancing the performance of the Deep Learning models used currently in the field; by harnessing the knowledge learnt from buildings with enough data, and use it to boost the accuracy of predictions for buildings with scarce data. To validate the approach, a case study for a large building was presented, where the data used in our experiments was representing only 2.5 months generated hourly. The paper compared the performance of the mostly used Deep Learning models in time-series forecasting before and after building-to-building transfer learning.

This work was the first in utilizing transfer learning and applying it to the Transformer model in the field of energy load forecasting, where the performance gain achieved by the Transformer according to MSE was 56.8% after applying our method. Future work will apply the same methodologies to different building domains and sizes to investigate the different outcomes. More research will focus into finding the limitations of this approach, mainly the least amount of data used with transfer learning to increase the performance.